\documentclass[]{interact}
\usepackage{epstopdf}
\usepackage{subfigure}
\usepackage{multirow}
\usepackage[backend=bibtex]{biblatex}
\usepackage{longtable}
\usepackage{caption}
\theoremstyle{plain}

\theoremstyle{definition}

\theoremstyle{remark}

\usepackage{xcolor}
\begin{document}

\articletype{ARTICLE TEMPLATE}

\title{GeoGPT: Understanding and Processing Geospatial Tasks through An Autonomous GPT}

\author{
\name{Yifan Zhang\textsuperscript{a, \#}, Cheng Wei\textsuperscript{a, \#},  Shangyou Wu\textsuperscript{a}, Zhengting He\textsuperscript{a}, and Wenhao Yu\textsuperscript{a, b, *}}
\affil{\textsuperscript{a} School of Geography and Information Engineering, China University of Geosciences, Wuhan, China; \textsuperscript{b} National Engineering Research Center for Geographic Information System, China University of Geosciences, Wuhan, China; \textsuperscript{\#} Authors with equal contribution to this paper; \textsuperscript{*} Corresponding author: Wenhao Yu (email: yuwh@cug.edu.cn)}
}

\maketitle

\begin{abstract}
    Generally, decision-makers in GIS need to combine a series of spatial algorithms and operations to solve geospatial tasks. For example, in the task of facility siting, the Buffer tool is usually first used to locate areas close or away from some specific entities; then, the Intersect or Erase tool is used to select candidate areas satisfied multiple requirements. Though professionals can easily understand and solve these geospatial tasks by sequentially utilizing relevant tools, it is difficult for non-professionals to handle these problems. Recently, Generative Pre-trained Transformer (e.g., ChatGPT) presents strong performance in semantic understanding and reasoning. Especially, AutoGPT can further extend the capabilities of large language models (LLMs) by automatically reasoning and calling externally defined tools. Inspired by these studies, we attempt to lower the threshold of non-professional users to solve geospatial tasks by integrating the semantic understanding ability inherent in LLMs with mature tools within the GIS community. Specifically, we develop a new framework called GeoGPT that can conduct geospatial data collection, processing, and analysis in an autonomous manner with the instruction of only natural language. In other words, GeoGPT is used to understand the demands of non-professional users merely based on input natural language descriptions, and then think, plan, and execute defined GIS tools to output final effective results. Several cases including geospatial data crawling, spatial query, facility siting, and mapping validate the effectiveness of our framework. Though limited cases are presented in this paper, GeoGPT can be further extended to various tasks by equipping with more GIS tools, and we think the paradigm of “foundational plus professional” implied in GeoGPT provides an effective way to develop next-generation GIS in this era of large foundation models.
\end{abstract}

\begin{keywords}
    Geospatial semantic understanding, AutoGPT, GeoAI, foundation model
\end{keywords}

\section{Introduction}
Since the advent of Geographical Information System (GIS), it has developed to be a powerful tool for many geospatial tasks, which provides researchers with the ability to collect, analyze, and visualize complex geospatial data [1], [2], [3]. In recent years, the integration of GIS in research has revolutionized the way we understand spatial relationships and presents strong assistance in various fields, such as urban planning [4], [5], environmental management [6], [7], and public health [8], [9]. To support these studies, numerous spatial algorithms and operations have been developed and designed, and many of these tools have been packed up and serve as the cornerstone of mainstream GIS software (e.g., ArcGIS and QGIS). Generally, to handle a geographical problem, researchers need to combine different GIS tools in sequence to solve different subgoals, respectively.

\begin{figure*}[h]
    \centering
    \includegraphics[width=0.99\textwidth]{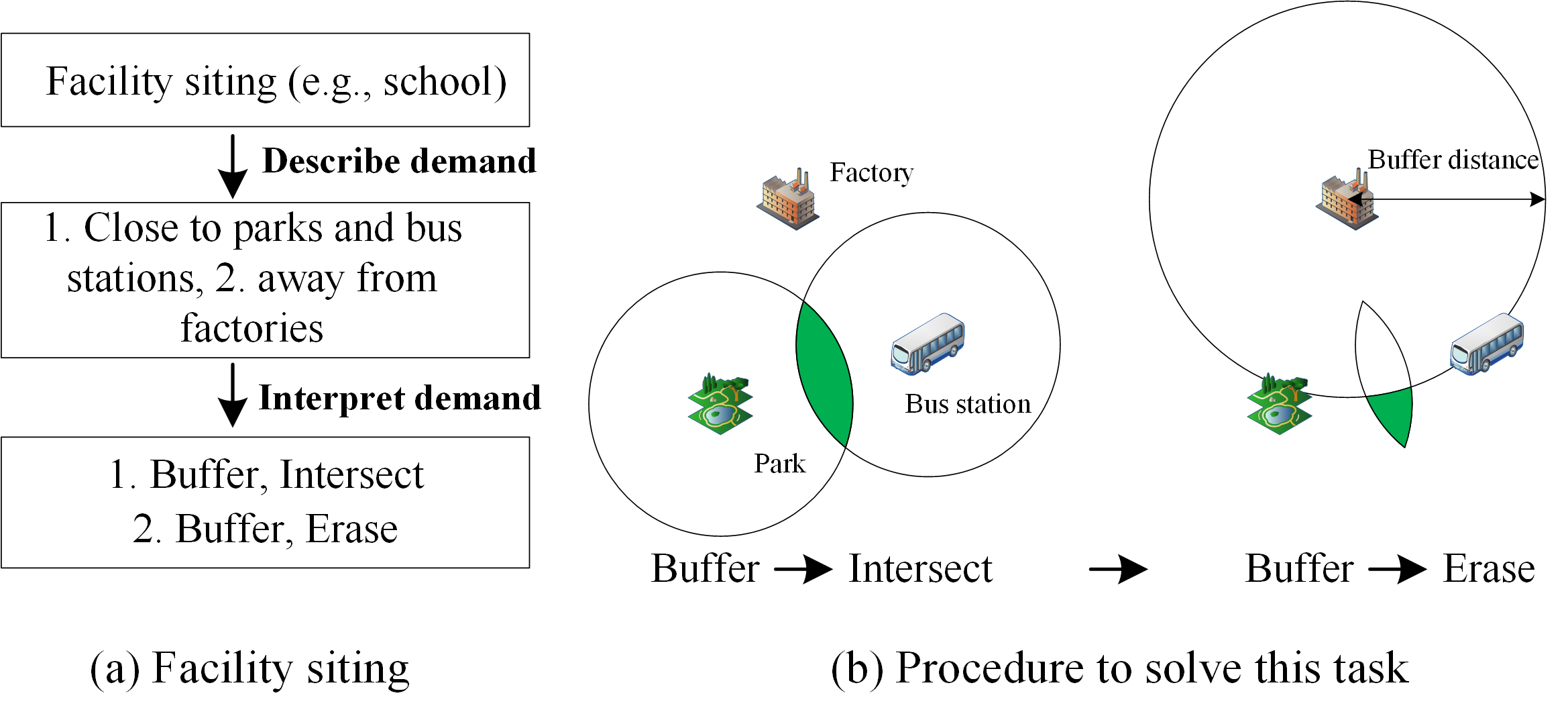} 
    \caption{A classical GIS task of facility siting. (a) Describe and interpret the demand of this task. (b) Procedure to solve this task by professionals.}
    \label{fig1}
    \end{figure*}

For example, considering a classical GIS task of facility siting in Figure 1, researchers tend to combine a series of operations to locate candidate areas that satisfied multiple requirements. Take school as the object of study, the traffic should usually be convenient and the environment should be pleasant, which means candidate locations of a school should be near public transport (e.g., subway or bus stations) and green space (e.g., parks or mountain); then, to avoid noise and environmental pollution, a school should be away from factories. To solve this problem, the Buffer tool is usually first used to find areas near transportation means and green space, then the Intersect tool is used to locate candidate areas near both the two types of entities; moreover, the Buffer tool is used to find areas near factories, then the Erase tool is used to remove these areas from candidate areas obtained from the last step, and the final result can be obtained (area of green in Figure 1b).

Therefore, it can be observed that though the aforementioned tools are already mature and the process can also be easily understood by professionals of GIS, there is still a large threshold for non-professionals to solve this problem. Some methods are proposed to ease this dilemma by first concatenating these tools in sequence by professionals and then providing them for non-professionals (e.g., ArcGIS ModelBuilder [10], [11]). In this way, non-professional users only need input required spatial datasets, and can then obtain final results. However, users can only use corresponding tools when they have entirely the same demands since these tools cannot be edited to be adaptive to different tasks after being released. Therefore, in this paper, we attempt to investigate a new framework that can autonomously and adaptively plan and execute GIS tools to solve geospatial tasks based on the diversified demands of non-professional users.

In this way, how to design frameworks or methods to accurately interpret the demand of non-professional users is a key problem. Generally, take the task of Figure 1 as an example, users may only describe the demand as “find places near parks and bus stations” rather than “use the Buffer tool on the layers of bus stations and parks to find places near them”. Moreover, as presented in Figure 1, automatically organizing accurate GIS tools in the right sequence based on multiple demands is also a challenge. 

In the past years, artificial intelligence especially large language models (LLMs) develop rapidly, which shows a strong potential to solve challenges in different fields [12], [13], [14]. Trained on very large-scale datasets, LLMs (e.g., ChatGPT) present surprising performances of semantic understanding and reasoning [15]. Moreover, recently proposed AutoGPT [16] can further extend the capabilities of LLMs by automatically reasoning and calling externally defined tools. Specifically, given a set of defined tools, AutoGPT employs LLMs as the agent or controller to understand demands, and then think and execute these tools step by step to solve complicated tasks. In this process, language is adopted as a generic interface to empower this. In this way, besides powerful language ability, LLMs can be equipped with professional skills. For example, though weak at precise numerical calculation, LLMs can finish this task by calling external calculation tools. Some pioneering research validates its effectiveness in various fields, such as computer vision [17], [18], chemistry [19], and public health [20].

Inspired by these studies, we attempt to lower the threshold of non-professional users to solve geospatial tasks by integrating the semantic understanding ability inherent in LLMs with mature tools within the GIS community. Specifically, we develop a new framework called GeoGPT that can conduct geospatial data collection, processing, and analysis in an autonomous manner with the instruction of only natural language. In our framework, some mature GIS tools are added, then an LLM is used to understand the demands of non-professional users merely based on input natural language descriptions. 
By thinking, planning, and executing these tools step by step, complicated geospatial tasks can be solved autonomously. Several classical cases including geospatial data crawling, spatial query, facility siting, and mapping validate the effectiveness of our framework. Based on these case studies, GeoGPT can understand what users needed from the perspective of GIS and accurately act to obtain final results. Though limited cases are presented in this paper, GeoGPT can be further extended to various tasks by equipping with more GIS tools. Finally, based on our investigation, we think the paradigm of “foundational plus professional” implied in GeoGPT provides an effective way to develop GeoAI in this era of large foundation models.

\section{Related work}
In recent years, the field of natural language processing (NLP) develops rapidly since the emergence of large language models (LLMs) [21], [22]. The representative models include GPT-3 [23], GPT-4 [24], PaLM [25], PaLM-2 [26], and LLaMa [27], to name a few. Equipped with large-scale models and training datasets, these LLMs present impressive zero-shot or few-shot performance in various tasks, such as semantic question answering [28], semantic reasoning [29], and even some complicated tasks beyond the field of NLP [30], [31]. For example, Chat-REC [32] introduces ChatGPT to augment the interactivity and explainability of an existing recommender system by converting user profiles and historical interactions into prompts. CaFo [33] takes advantage of abundant knowledge implied in LLMs for generating more concrete descriptions of objects of interest, which can then be used to assist image generation tools (e.g., DALL-E [34], [35]) to output better images. In the field of GIS, researchers have also investigated whether LLMs can understand and then solve geospatial tasks [36]. For example, [37] tests the spatial semantic reasoning ability of ChatGPT on geospatial tasks including toponym recognition, location description recognition, and US state-level/county-level dementia time series forecasting, which suggests that ChatGPT can outperform task-specific fully-supervised models in a zero-shot or few-shot learning setting.

However, though LLMs have demonstrated impressive performance across tasks in language understanding, solving complicated tasks within different domains based on mere LLMs remains a key challenge [38], [39]. Especially, tasks in some professional fields (such as the cartography task in GIS) are complicated and generally rely on the utilization of multiple professional tools or procedures. Therefore, to extend the ability of LLMs to other professional fields, some pioneering studies attempt to convert the role of LLMs from executive to decision-maker [40]. In other words, instead of directly solving tasks, the exceptional ability of LLMs in language understanding, interaction, and reasoning is used for analyzing demands and selecting professional tools, and tasks in different domains will then be solved by chosen tools. After some previous attempts [41], [42], [43], [44], [45], [46], a representative method called ReAct [47] proposes a feasible and relatively complete framework. In this framework, LLMs can be used for synergizing reasoning and acting, and the process is described as the iteration of thought (i.e., semantic understanding and reasoning), action (i.e, use of professional tools), observation (i.e., obtaining results of actions), and thought again based on observations and initial target.

In this way, the ability of LLMs to solve professional tasks can be greatly enhanced with the equipment of external tools. Some research from different fields validates its effectiveness. For example, AutoGPT can be directed to crawl external information through a wide array of online platforms. Visual ChatGPT [17] combines ChatGPT and various external visual foundation models, which can solve complex visual questions step-by-step. Then, HuggingGPT [48] further extends LLMs with various AI models in machine learning communities (e.g., Hugging Face), which can solve complicated AI tasks with different domains and modalities. Besides the field of computer science, some researchers from other fields also show the effectiveness of integrating LLMs with professional tools to automatically solve complicated tasks, such as organic synthesis and drug discovery in chemistry [19] and infodemiology in public health [20]. Inspired by these studies, we propose a framework called GeoGPT that combines LLMs with mature tools within the GIS community to solve various geospatial tasks. Our framework can benefit non-professional users since it can accept natural language input based on the strong semantic understanding performance of LLMs.

\section{GeoGPT}
\subsection{Framework}
In this section, we will introduce GeoGPT, an LLM-based framework we developed to automatically solve geospatial tasks by a chain of thinking mode based on the demands of non-professional users. GeoGPT uses the Langchain [49] framework to connect an LLM (we use gpt-3.5-turbo with a temperature of 0 in our experiments) with various mature tools within the GIS community. In this way, GeoGPT can automatically choose appropriate GIS tools for geographical dataset collection, spatial data analysis, and mapping, to name a few. The overall framework of GeoGPT is shown in Figure 2. In GeoGPT, we construct a GIS tool pool that contains various GIS tools to support geospatial applications.

\begin{figure*}[h]
    \centering
    \includegraphics[width=0.99\textwidth]{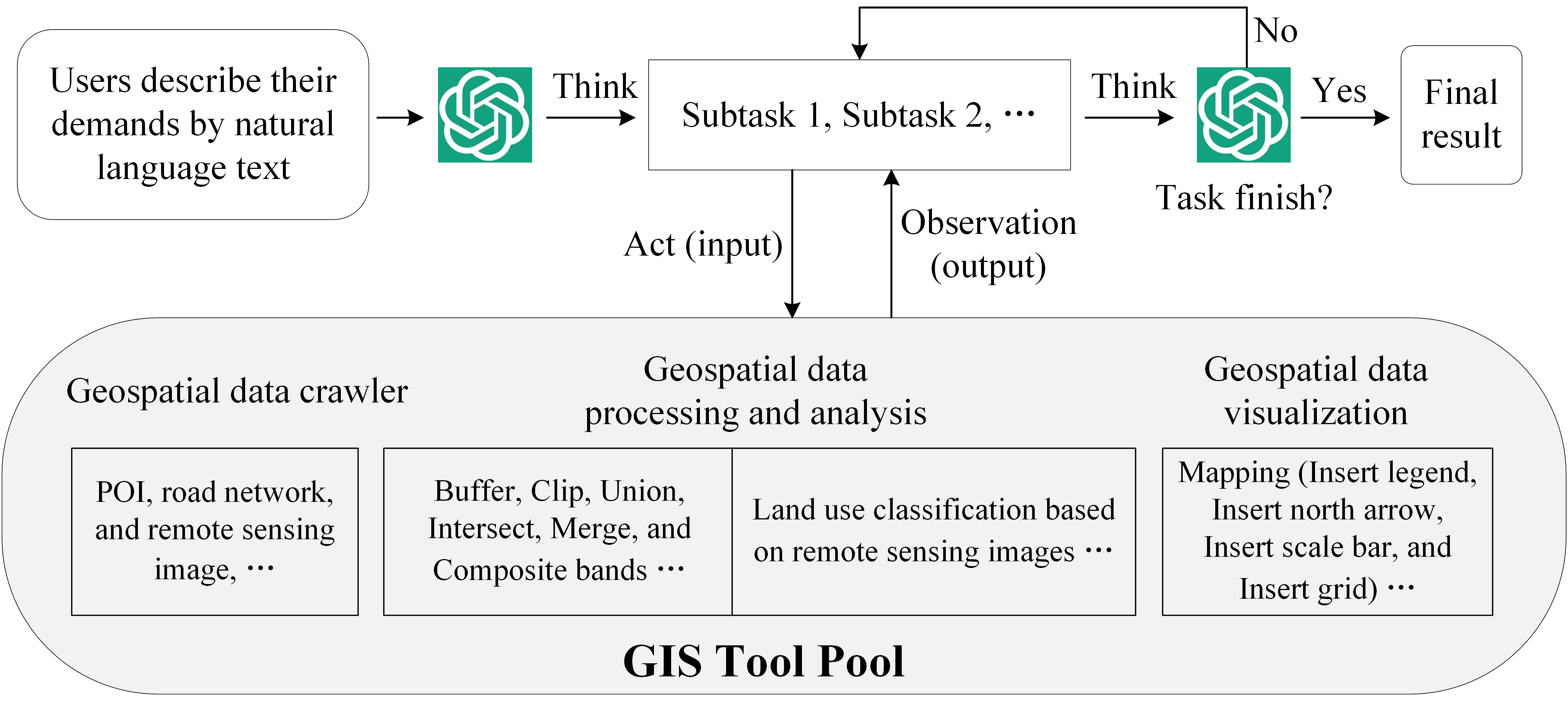} 
    \caption{The framework of our GeoGPT. Users first describe their demands by natural language text, then an LLM is used for interpreting and reasoning demands implied in the text, and chooses and calls appropriate tools in GIS Tool Pool to solve geospatial tasks step by step. The final result can be obtained when the LLM thinks the task is finished.}
    \label{fig2}
    \end{figure*}

\subsection{LangChain}
LangChain is a comprehensive framework designed to extend the ability of LLMs and facilitate the development of language model applications [49]. In this framework, various modules including access to language models, prompts input, document loaders, chains, indexes, agents, memory, and chat functionality are defined, and researchers can easily integrate these modules for their personalized demands, such as chatbots and question answering systems. Moreover, LangChain also provides the interface to use external tools, which means users can integrate both strong abilities of semantic understanding and reasoning of LLMs and professional tools in different fields. Generally, the prompt presented in Figure 3 is adapted to guide LLMs to identify and use appropriate external tools to solve professional tasks step by step [20]. In this paper, we also use this prompt in GeoGPT.

\begin{figure*}[h]
    \centering
    \includegraphics[width=0.99\textwidth]{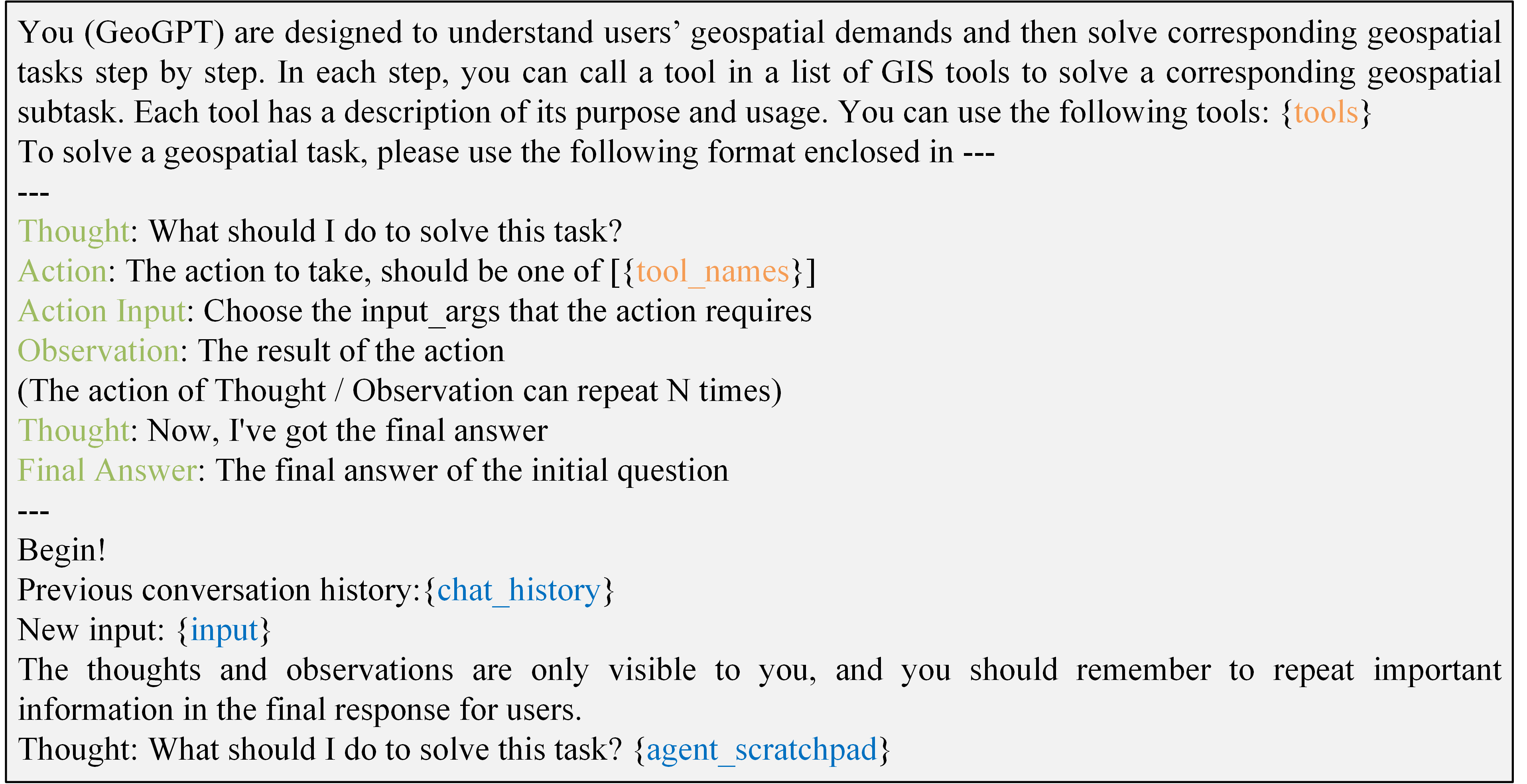} 
    \caption{The prompt used in our experiments to specifies the format of GeoGPT to understand and process geospatial tasks. In this prompt, GeoGPT is guided to call different GIS tools to solve corresponding geospatial tasks based on the understanding of both users’ demands and tools’ descriptions.}
    \label{fig3}
    \end{figure*}

\subsection{Tools}
Over the last few decades, within the GIS community, numerous spatial algorithms and operations have been developed and designed to handle a wide range of geospatial tasks. In this paper, to test the ability of GeoGPT to solve geospatial tasks, we design various GIS tools from three aspects including geospatial data collection, processing and analysis, and visualization, as listed in Table 1. It should be noted the tools listed here are merely for validating the effectiveness of our framework on some classical geospatial tasks, and GeoGPT can have more personalized functions for different demands, as it can be easily extended by designing more tools.

\begin{longtable}{|p{6cm}|p{8cm}|}
\caption{Tools implemented in GeoGPT.}
\label{table1}
\\
\hline
\textbf{Tools} & \textbf{Description} \\
\hline
\endfirsthead
\hline
\endhead

\hline
\multicolumn{2}{r}{\textit{Continued on next page}} \\
\endfoot

\hline
\endlastfoot

\multicolumn{2}{|c|}{\textbf{Data Collection Tools}}  \\
\hline
Get\_POI\_By\_Keywords & Use this tool when you need to get POI data for a Chinese city by POI keywords. The input of this tool must at least provide the city name, after which the user can provide multiple POI categories or none. City names and each POI category are separated by commas. \\
\hline
Get\_POI\_By\_Polygon & Use this tool when you need to get POI data within a polygonal area. Input a longitude-latitude coordinate pair representing the area of the polygon. Longitude and latitude are separated by ",", and the coordinate pairs are separated by "$\mid$ ". The first and last coordinate pairs need to be the same. \\
\hline
Get\_Road\_Network\_By\_Rectangle & Use this tool when you need to get road network data within a rectangular area. Input a comma-separated string of four-tuples, representing two longitude ranges x1, x2 enclosed in a rectangle, and two latitude ranges y1, y2 enclosed in a rectangle. \\
\hline
Get\_Remote\_Sensing\_Image & Input parameters, in order of province, city, start time and end time, in comma-separated strings \\
\hline
\multicolumn{2}{|c|}{\textbf{Data Loader Tools (Load data locally)}}  \\
\hline
Load\_Subway\_Data & Load the dataset of subway stations, return the file path of subway stations. \\
\hline
Load\_Hotel\_Data & Load the dataset of hotels, return the file path of hotels. \\
\hline
Load\_Factory\_Data & Load the dataset of factories, return the file path of factories. \\
\hline
Load\_Supermarket\_Data & Load the dataset of supermarkets, return the file path of supermarkets. \\
\hline
Load\_Remote\_Sensing\_Image\_Data & Load the dataset of remote sensing image, return the file path of RS image. \\
\hline
Load\_Wuhan\_Main\_Urban\_Data & Load the dataset of Wuhan main urban, return the file path of Wuhan main urban. \\
\hline
\multicolumn{2}{|c|}{\textbf{Spatial Analysis Tools}} \\
\hline
Obtain location & Input a place name, return its location. Use it when you need to find unknown locations. \\
\hline
Buffer & Input "shapefile, number", return the neighboring range around this shapefile. \\
\hline
Intersect & Input two shapefiles, retain objects meanwhile within (inside) both shapefiles. \\
\hline
Clip & Input two shapefiles "A, B", return (keep) objects of A only within (inside) the range of B. \\
\hline
Erase & Input two shapefiles "A, B", remove (erase) objects of A only within (inside) the range of B. \\
\hline
\multicolumn{2}{|c|}{\textbf{Tools for RS Image Processing} } \\
\hline
Crop\_Remote\_Sensing\_Image & Use this tool when you need to crop an image of Remote Sensing. Input a tif file storing remote sensing and a shapefile storing the cropping area, which are separated by a comma. \\
\hline
LandUse\_Classification & Use this tool when you need to perform land use classification on remote sensing imagery in a tif file. Input a tif file storing remote sensing image, and output a raster file (tif format) after land use classification. \\
\hline
Raster\_to\_Vector & Use this tool when you need to convert raster files (e.g., tif format) to vector files. Input a raster file, and output the storage path of the converted vector file. \\
\hline
\multicolumn{2}{|c|}{\textbf{Cartography Tool}} \\
\hline
Mapping & Use this tool when you need to draw a map using geographic data stored in vector files (e.g., shapefiles). The input must be a vector file that needs to draw a map. The tool outputs the storage path of the drawn map. \\
\hline

\end{longtable}

\subsubsection{Geospatial data collection}
In GIS, collecting effective geospatial data is a crucial premise for various geospatial tasks. Nowadays, the crowdsource data platform and the crawler tool facilitate the acquisition of a wide range of geographic data according to user needs. Therefore, in GeoGPT, we expect that the input data can be defined by users or crawled from the Internet according to user needs. In this way, we design tools for collecting three representative datasets including point of interest (POI), road network, and remote sensing image. 

\paragraph{POI Collection}
We define a crawler function to obtain POI datasets from the internet. To make GeoGPT friendly to non-professional users, GeoGPT can accept natural language text input. For example, users can describe their demand as “download POI in the city of Beijing” or “download POI of the restaurant in the city of Beijing”, and GeoGPT will first analyze this demand via semantic understanding and then call POI Collection tool by inputting keywords of “Beijing” or “restaurant in the city of Beijing” to obtain corresponding results. Moreover, users can also download POI in areas with specific shapes, such as “given an enclosed area described by continuous pairs of latitude and longitude: (39.820, 116.260), (39.990, 116.260), (39.990, 116.490), (39.820, 116.490), download POI in this area”.

\paragraph{Road network Collection}
We define a function to download road networks from OpenStreetMap (https://www.openstreetmap.org/) to obtain road network datasets. Similarly, users can use similar text commands to that in POI Collection to download road networks 

\paragraph{Remote sensing image Collection}
We define a function to download remote sensing images from the website of Geospatial Data Cloud (https://www.gscloud.cn/). Similarly, users can use similar text commands to that in POI Collection for remote sensing image collection. Users can download remote sensing images including LANDSAT, MODIS, and DEM, to name a few. 

\subsubsection{Geospatial data processing and analysis}
To support the ability of spatial analysis and spatial data mining, we also add tools for geospatial data processing and analysis into GeoGPT. Specifically, some traditional commonly used GIS tools are added including \textbf{ Buffer}, \textbf{Clip}, \textbf{Union}, \textbf{Intersect}, \textbf{Erase}, and \textbf{Composite bands}, to name a few. Then, we also add some tools for specific applications, such as \textbf{Land use classification based on remote sensing images}. Applications of these tools will be illustrated in case studies.

\subsubsection{Geospatial data visualization}
In GIS, geospatial data visualization (\textbf{Mapping}) is also a key part for users to obtain information. Therefore, we also provide some cartographic tools in GeoGPT, such as \textbf{Insert legend}, \textbf{Insert north arrow}, \textbf{Insert scale bar}, and \textbf{Insert gird}. With these tools, users can be provided maps with corresponding elements.

\section{Case studies}
In this section, our GeoGPT is applied to several classical case studies in the GIS field, including geospatial data download, facility sitting, spatial query, and thematic mapping. Details of the input and output of GeoGPT will be described and discussed to further show how it works.
\subsection{Geospatial data crawling}
Geospatial data acquisition is one of the key capabilities of GeoGPT. In order to validate the wide range of GeoGPT data acquisition, we designed several data download queries for the download requirements of the user on POI data, road network data, and remote sensing image data. As shown in Tables 2-5, GeoGPT comprehends the data download requirements specified in the query and matches them with the appropriate descriptions of the tools, invoking the relevant tools to fulfill the specific data download needs. 

\begin{longtable}{|p{14cm}|}
    \caption{The case study of data download for POI by keywords.}
    \label{table2}
    \\
    \hline
    \textbf{Query}: Download POI of the restaurants and hotels in the city of Beijing. \\
    \hline
    \textbf{Thought}: \\
    \textbf{Action}: Get\_POI\_By\_Keywords\\
    \textbf{Action Input}: Beijing, restaurants, hotels\\
    \textbf{Observation}: In this step I obtain POIs that meet the requirements, it is stored at ./data\_output/POI/POI.csv\\
\\
    \textbf{Thought}: \\
    \textbf{Action}: None\\
\\
    \textbf{Final Answer}: The POI data for restaurants and hotels in the city of Beijing can be found in the file ./data\_output/POI/POI.csv.\\
    \hline
    \end{longtable}

    \begin{longtable}{|p{14cm}|}
        \caption{The case study of data download for POI by polygonal area.}
        \label{table3}
        \\
        \hline
        \textbf{Query}: Given an enclosed area described by continuous pairs of latitude and longitude: (39.820, 116.260), (39.990, 116.260), (39.990, 116.490), (39.820, 116.490), download POI in this area \\
        \hline
        \textbf{Thought}:\\
        \textbf{Action}: Get\_POI\_By\_Polygon\\
        \textbf{Action Input}: "39.820,116.260$\mid$39.990,116.260$\mid$39.990,116.490$\mid$39.820,116.490"\\
        \textbf{Observation}: In this step I obtain POIs that meet the requirements, it is stored at ./data\_output/POI/POI.csv\\
\\
        \textbf{Thought}: Review original question and check my total process\\
\\
\textbf{Final Answer}: The POI data for the enclosed area described by the continuous pairs of latitude and longitude (39.820, 116.260), (39.990, 116.260), (39.990, 116.490), (39.820, 116.490) can be found in the file ./data\_output/POI/POI.csv.\\

        \hline
        \end{longtable}

        \begin{longtable}{|p{14cm}|}
            \caption{The case study of data download for road network by rectangular area.}
            \label{table4}
    \\
            \hline
            \textbf{Query}:  Download the road network data for the rectangular area with a longitude range of -8.7 to -8.519847 and a latitude range of 41.064999 to 41.189018 \\
            \hline
            \textbf{Thought}: \\
            \textbf{Action}: Get\_Road\_Network\_By\_Rectangle\\
            \textbf{Action Input}: -8.7,-8.519847,41.064999,41.189018\\
            \textbf{Observation}: Spend time: 34.473637342453 seconds. In this step I obtain Road Networks that meet the requirements, it is stored at ./data\_output/road\_network\\
\\
            \textbf{Thought}: Review original question and check my total process\\
\\
            \textbf{Final Answer}: The road network data for the rectangular area with a longitude range of -8.7 to -8.519847 and a latitude range of 41.064999 to 41.189018 is stored at ./data\_output/road\_network.\\

            \hline
        \end{longtable}
    
        \begin{longtable}{|p{14cm}|}
            \caption{The case study of data download for remote sensing images.}.
            \label{table5}
    \\
            \hline
            \textbf{Query}: Download remote sensing images of Wuhan City, Hubei Province from February 10, 2015 to February 20, 2015. \\
            \hline
            \textbf{Thought}: I need to download remote sensing images of Wuhan City, Hubei Province from February 10, 2015 to February 20, 2015.\\
            \textbf{Action}: Get\_Remote\_Sensing\_Image\\
            \textbf{Action Input}: "Hubei, Wuhan, 2015-02-10, 2015-02-20"\\
            Please enter verification code:YXNF\\
            Start downloading data ……\\
            \textbf{Observation}: Remote sensing data has been downloaded and saved in ./data\_output/rs\_image. My work is done!\\
\\
            \textbf{Thought}:\\
\\
            \textbf{Final Answer}: Remote sensing images of Wuhan City, Hubei Province from February 10, 2015 to February 20, 2015 have been downloaded and saved in ./data\_output/rs\_image.\\

            \hline
        \end{longtable}

\subsection{Facility sitting}
To test the ability of GeoGPT, we use GeoGPT to solve facility sitting, which is a classical spatial analysis task in GIS. Given an example query in Table 6, three demands are given to describe the candidate area. In this case, we prepare three existing datasets with corresponding data load tools, and GeoGPT should determine when and which dataset it should use. From Table 6, it can be observed that GeoGPT first loads the datasets needed, and then use the Buffer tool to obtain their neighboring areas. Then, it takes the Erase tool to erase the neighboring areas of factories from neighboring areas of subway stations. Finally, it takes the Intersect tool to obtain candidate areas by intersecting neighboring areas of supermarkets and the result from the last step. Based on this case, we can observe that GeoGPT can plan a sequential process, and it can also remember to use previous datasets it has obtained.

\begin{longtable}{|p{14cm}|}
    \caption{The case study of facility sitting.}.
    \label{table6}
\\

    \hline
    \textbf{Query}: I want to find areas that locate meanwhile within a maximum distance of 500 meters from the subway stations, within neighboring areas of 800 meters from the supermarkets, and not within neighboring areas of 1500 meters from the factories. \\
    \hline
    \endfirsthead
    \hline
    \endhead

\hline
\multicolumn{1}{r}{\textit{Continued on next page}} \\
\endfoot

\hline
\endlastfoot
\hline

    \textbf{Thought}: I need to find areas that locate within a maximum distance of 500 meters from the subway stations, within neighboring areas of 800 meters from the supermarkets, and not within neighboring areas of 1500 meters from the factories.\\
    \textbf{Action}: Load\_Subway\_Data\\
    \textbf{Action Input}: None\\
    \textbf{Observation}: I obtain the dataset of subway stations, it is stored in ./data\_use/case2/subway\_station.shp\\
    \\
    \textbf{Thought}: I need to buffer the subway stations dataset to get the areas within a maximum distance of 500 meters from the subway stations.\\
    \textbf{Action}: Buffer\\
    \textbf{Action Input}:  ./data\_use/case2/subway\_station.shp, 500\\
    \textbf{Observation}: I obtain the neighboring area with a distance of 500.0 for ./data\_use/case2/subway\_station.shp, it is stored in ./data\_output/subway\_station\_neighborhood\_500.0.shp\\
    \\
    \textbf{Thought}: I need to load the dataset of supermarkets.\\
    \textbf{Action}: Load\_Supermarket\_Data\\
    \textbf{Action Input}: None\\
    \textbf{Observation}: I obtain the dataset of supermarkets, it is stored in ./data\_use/case2/supermarket.shp\\
    \\
    \textbf{Thought}: I need to buffer the supermarkets dataset to get the areas within a neighboring area of 800 meters from the supermarkets.\\
    \textbf{Action}: Buffer\\
    \textbf{Action Input}: ./data\_use/case2/supermarket.shp, 800\\
    \textbf{Observation}: I obtain the neighboring area with a distance of 800.0 for ./data\_use/case2/supermarket.shp, it is stored in ./data\_output/supermarket\_neighborhood\_800.0.shp\\
    \\
    \textbf{Thought}: I need to load the dataset of factories.\\
    \textbf{Action}: Load\_Factory\_Data\\
    \textbf{Action Input}: None\\
    \textbf{Observation}: I obtain the dataset of factories, it is stored in ./data\_use/case2/factory.shp\\
    \\
    \textbf{Thought}: I need to buffer the factories dataset to get the areas not within a neighboring area of 1500 meters from the factories.\\
    \textbf{Action}: Buffer\\
    \textbf{Action Input}: ./data\_use/case2/factory.shp, 1500\\
    \textbf{Observation}: I obtain the neighboring area with a distance of 1500.0 for ./data\_use/case2/factory.shp, it is stored in ./data\_output/factory\_neighborhood\_1500.0.shp\\
    \\
    \textbf{Thought}: I need to erase the areas not within a neighboring area of 1500 meters from the factories from the areas within a maximum distance of 500 meters from the subway stations and within neighboring areas of 800 meters from the supermarkets.\\
    \textbf{Action}: Erase\\
    \textbf{Action Input}: ./data\_output/subway\_station\_neighborhood\_500.0.shp, ./data\_output/factory\_neighborhood\_1500.0.shp\\
    \textbf{Observation}: In this step I obtain objects that locate within ./data\_output/subway\_station\_neighborhood\_500.0.shp but not in ./data\_output/factory\_neighborhood\_1500.0.shp, it is stored in ./data\_output/erase.shp\\
    \\
    \textbf{Thought}: I need to intersect the areas within a maximum distance of 500 meters from the subway stations and within neighboring areas of 800 meters from the supermarkets.\\
    \textbf{Action}: Intersect\\
    \textbf{Action Input}: ./data\_output/erase.shp,\\ ./data\_output/supermarket\_neighborhood\_800.0.shp\\
    \textbf{Observation}: In this step I obtain objects that meanwhile locate within ./data\_output/supermarket\_neighborhood\_800.0.shp,./data\_output/erase.shp, it is stored in ./data\_output/intersect.shp\\
    \\
    \textbf{Final Answer}: The areas that locate meanwhile within a maximum distance of 500 meters from the subway stations, within neighboring areas of 800 meters from the supermarkets, not within neighboring areas of 1500 meters from the factories can be found in the file ./data\_output/intersect.shp.\\
    
    \hline
\end{longtable}

\subsection{Spatial query}
In this section, we also test GeoGPT on the task of spatial query. In effect, this task is similar but different from facility sitting, which is implemented to test whether GeoGPT can distinguish a specific dataset of “hotels” and the “areas” in facility sitting. From the result in Table 7, it can be observed that GeoGPT can successfully interpret the demand and finish this task by first using the Buffer tool for two datasets (subway stations and supermarkets), and then using the Intersect tool to obtain areas within both of them, and finally using the Clip tool to obtain hotels within these areas.

\begin{longtable}{|p{14cm}|}
    \caption{The case study of spatial query.}.
    \label{table7}
\\

    \hline
    \textbf{Query}: I want to find hotels that locate meanwhile within a maximum distance of 500 meters from subway stations, and within neighboring areas of 800 meters from supermarkets. \\
    \hline
    \endfirsthead
    \hline
    \endhead

\hline
\multicolumn{1}{r}{\textit{Continued on next page}} \\
\endfoot

\hline
\endlastfoot
\textbf{Thought}: I need to find hotels that are within 500 meters of subway stations and 800 meters of supermarkets. I need to use the tools provided to find the answer.\\
\textbf{Action}: Load\_Hotel\_Data\\
\textbf{Action Input}: None\\
\textbf{Observation}: I obtain the dataset of hotels, it is stored in ./data\_use/case3/hotel.shp\\
\\
\textbf{Thought}: I need to find the locations of subway stations and supermarkets.\\
\textbf{Action}: Load\_Subway\_Data\\
\textbf{Action Input}: None\\
\textbf{Observation}: I obtain the dataset of subway stations, it is stored in ./data\_use/case3/subway\_station.shp\\
\\
\textbf{Thought}:\\
\textbf{Action}: Load\_Supermarket\_Data\\
\textbf{Action Input}: None\\
\textbf{Observation}: I obtain the dataset of supermarkets, it is stored in ./data\_use/case3/supermarket.shp\\
\\
\textbf{Thought}:\\
\textbf{Action}: Buffer\\
\textbf{Action Input}: ./data\_use/case3/subway\_station.shp, 500\\
\textbf{Observation}: I obtain the neighboring area with a distance of 500.0 for ./data\_use/case3/subway\_station.shp, it is stored in ./data\_output/subway\_station\_neighborhood\_500.0.shp\\
\\
\textbf{Thought}:\\
\textbf{Action}: Buffer\\
\textbf{Action Input}: ./data\_use/case3/supermarket.shp, 800\\
\textbf{Observation}: I obtain the neighboring area with a distance of 800.0 for ./data\_use/case3/supermarket.shp, it is stored in ./data\_output/supermarket\_neighborhood\_800.0.shp\\
\\
\textbf{Thought}:\\
\textbf{Action}: Intersect\\
\textbf{Action Input}: ./data\_output/subway\_station\_neighborhood\_500.0.shp, ./data\_output/supermarket\_neighborhood\_800.0.shp\\
\textbf{Observation}: In this step I obtain objects that meanwhile locate within ./data\_output/supermarket\_neighborhood\_800.0.shp,\\./data\_output/subway\_station\_neighborhood\_500.0.shp, it is stored in ./data\_output/intersect.shp\\
\\
\textbf{Thought}:\\
\textbf{Action}: Clip\\
\textbf{Action Input}: ./data\_use/case3/hotel.shp, ./data\_output/intersect.shp\\
\textbf{Observation}: In this step I obtain objects of ./data\_use/case3/hotel.shp within ./data\_output/intersect.shp, it is stored in ./data\_output/clip.shp\\
\\
\textbf{Final Answer}: The hotels that locate meanwhile within a maximum distance of 500 meters from subway stations, and within neighboring areas of 800 meters from supermarkets are stored in ./data\_output/clip.shp.\\

\hline

    \hline
\end{longtable}

\subsection{Mapping}
In this case, we test the ability of GeoGPT to meet the requirements of mapping. As shown in Table 8, we want a thematic map of land use in the main urban area of Wuhan. Therefore, it is necessary to cut out the remote sensing image corresponding to the main urban area of Wuhan from the original remote sensing image on a small scale, and use relevant tools to classify the land use and map it. GeoGPT makes it easy to understand the requirements in the query as well as the GIS operation logic, and then invokes the target tool step by step to implement it. Corresponding steps are presented in Figure 4.

\begin{longtable}{|p{14cm}|}
    \caption{The case study of mapping.}.
    \label{table8}
\\

    \hline
    \textbf{Query}:I want a visual map of cropped remote sensing image based on the main urban area of Wuhan, whose land use has been classified. \\
    \hline
    \endfirsthead
    \hline
    \endhead

\hline
\multicolumn{1}{r}{\textit{Continued on next page}} \\
\endfoot

\hline
\endlastfoot
\textbf{Thought}: I need to first load the datasets of remote sensing image and Wuhan main urban, then crop the remote sensing image based on the main urban area, then perform land use classification on the cropped image, and finally draw a map using the classified image.\\
\textbf{Action}: Load\_Remote\_Sensing\_Image\_Data\\
\textbf{Action Input}: None\\
\textbf{Observation}: I obtain the dataset of remote sensing image, it is stored in ./data\_use/case4/data.tif\\
\\
\textbf{Thought}: I need to load the dataset of Wuhan main urban.\\
\textbf{Action}: Load\_Wuhan\_Main\_Urban\_Data\\
\textbf{Action Input}: None\\
\textbf{Observation}: I obtain the dataset of Wuhan main urban, it is stored in  ./data\_use/case4/WuhanMainCountry.shp\\
\\
\textbf{Thought}: I need to crop the RS image using the Wuhan main urban as the cropping area.\\
\textbf{Action}: Crop\_Remote\_Sensing\_Image\\
\textbf{Action Input}: ./data\_use/case4/data.tif, ./data\_use/case4/WuhanMainCountry.shp\\
\textbf{Observation}: In this step I obtain cropped RS image that meet the requirements, it is stored at ./data\_output/RS\_crop.tif\\
\\
\textbf{Thought}: I need to perform land use classification on the cropped RS image.\\
\textbf{Action}: LandUse\_Classification\\
\textbf{Action Input}: ./data\_output/RS\_crop.tif\\
\textbf{Observation}: In this step I obtain RS image after land use classification, it is stored in a raster file at ./data\_output/lucc.tif\\
\\
\textbf{Thought}: I need to draw a map using the classified RS image.\\
\textbf{Action}: Mapping\\
\textbf{Action Input}: ./data\_output/lucc.tif\\
\textbf{Observation}: An exception occurred using this tool, I need to check whether the input file is a vector file, whether the input file is exists, and whether I have selected the wrong tool.\\
\\
\textbf{Thought}: \\
\textbf{Action}: Raster\_to\_Vector\\
\textbf{Action Input}: ./data\_output/lucc.tif\\
\textbf{Observation}: In this step I obtain the converted vector file (shapefile format) based on the raster file, it is stored at ./data\_output/polyone.shp\\
\\
\textbf{Thought}: I need to draw a map using the vector file.\\
\textbf{Action}: Mapping\\
\textbf{Action Input}: ./data\_output/polyone.shp\\
\textbf{Observation}: In this step I obtain the visual map that meet the requirements, it is stored at ./data\_output/visual\_map.png\\
\\
\textbf{Final Answer}: The visual map of cropped remote sensing image based on the Wuhan main urban, whose land use has been classified, is stored at ./data\_output/visual\_map.png.\\

\hline

    \hline
\end{longtable}

\begin{figure*}[htbp]
    \centering
    \includegraphics[width=0.99\textwidth]{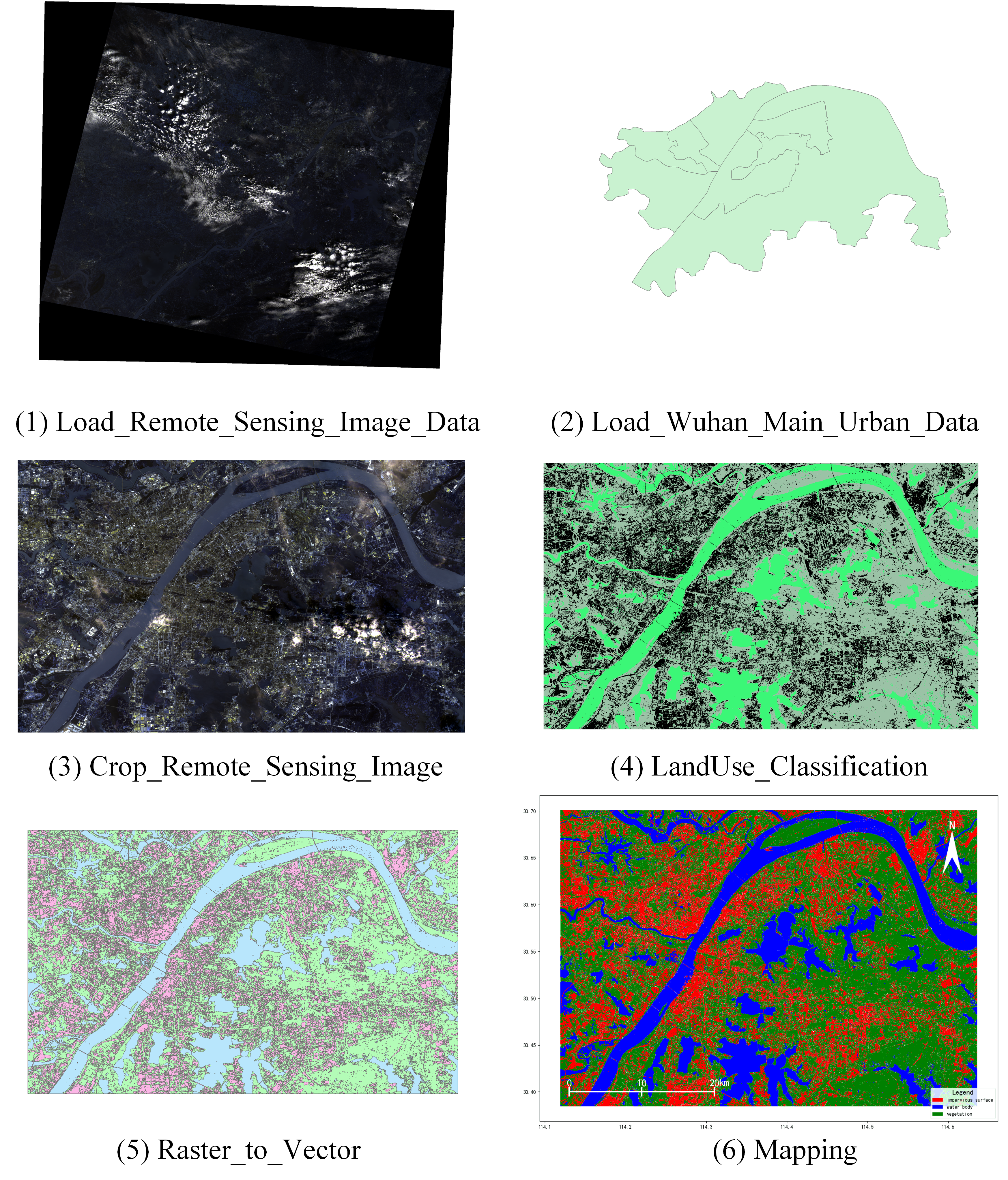} 
    \caption{The visualization of steps presented in Table 8.}
    \label{fig4}
    \end{figure*}

\section{Discussion}
\subsection{Uncertainty}
First, the uncertainty of GeoGPT is discussed. In our case studies, the LLM of gpt-3.5-turbo is used as the agent. In effect, the uncertainty of the LLM output will influence our results, since the analysis results in GIS should always be stable and accurate. During the testing process, different results or even failures may appear with the same experimental settings and codes. Moreover, the LLM seems to be sensitive to some words. For example, if we add (or remove) “the”, “and”, “but”, or “1”, “2” into (or from) the query, the results may sometimes also be different. In our case studies, we try our best to avoid such situations including adding prompts, adjusting tool descriptions, setting the temperature of the LLM, and designing protection mechanisms in tools (Section 5.3). However, though these measures can relieve this problem, it still exists. Similar problems are also found in other relevant studies. Therefore, we think stabilizing the output of LLMs is a direction worth exploring in professional fields.

\subsection{Language understanding}
It is easy for professionals in the field of GIS to understand the functions of different tools. Since the training datasets of LLMs contain knowledge from various fields, it also knows a lot about GIS. For example, if you ask ChatGPT the question “Do you understand the GIS tool Clip”, you will obtain an accurate answer about this tool “The Clip tool is used to extract or isolate a subset of geographic data based on a defined boundary or extent”, which means you can obtain a subset of a global geographic dataset in such a defined boundary and objects outside this boundary will be removed. In the GIS tool “Clip”, the input contains two spatial datasets including one for global geographic data and one for a boundary.

However, there is also a similar but different meaning of “Clip” in LLMs. For example, in the sentence “I need to clip some photos from the magazine for my art project”, users want to retain photos; but in the sentence “I need to clip the grass in the garden to make it look neat and tidy”, users want to remove grass. Therefore, our defined tool “Clip” may sometimes conflict with other meanings of “clip” inherent in LLMs. Since the input sequence (dataset and boundary) of our defined “Clip” tool is important and determines clip results, such conflict may bring confusion to LLMs and then output undesired results. To solve this problem, we describe the tool “Clip” as “Input two shapefiles "A, B", return (keep) objects of A only within (inside) the range of B”. In the description, we use “\textbf{keep} objects in B” rather than “\textbf{remove} objects outside B” to reduce the gap between our tool description and original knowledge in LLMs. However, it is also limited and GeoGPT may sometimes be confused about the usage of “Clip”, “Erase”, and “Intersect”, such a condition may be worse with more tools. Therefore, we think investigating a feasible solution to align and harmonize professional descriptions and universal knowledge in LLMs is a direction worth studying.

\subsection{Protection mechanism}
Since the uncertainty of LLMs has an impact on the results of GeoGPT, we attempt to relieve such conditions by designing protection mechanisms. Take Clip as an example, the input should be a string containing file paths of two datasets, and the former is used as the global dataset while the latter is used as the clip boundary. Therefore, there is a strict constraint when using “Clip”, which means the clip boundary should have higher dimension geometry than the global dataset. For example, when dimension of the global dataset is polygon, the clip boundary must also be polygon; when dimension of the global dataset is line, the clip boundary should be line or polygon. In this way, in the Clip tool, a pre-judgment will be conducted to confirm whether the input satisfies this requirement. If not, the output (Observation) will return “An exception occurred using this tool, I need to check whether the input file exists, or whether the input contains excessive content, or whether the input format is correct, or whether the input file order is correct, or and whether I have selected the wrong tool” to let GeoGPT check related contents. This protection mechanism is also effective when there are some hallucinated files and tools. For example, in example (1) in Table 9, GeoGPT needs to use Buffer for the dataset of subway stations, but it inputs only the name of “Subway Stations” rather than the corresponding file path. Therefore, the tool will return such exception information to remind GeoGPT to check the input, and GeoGPT then knows to use Load\_Subway\_Data to obtain subway stations. A similar problem also appears in example (2). Then, in example (3), GeoGPT chooses a hallucinated tool Load\_Hubei\_University\_Data to obtain the location of “Hubei University”, and the tool is practically non-existent in our tool list. During our whole test stage, we find that LLMs can imitate our code style and create some tools and file paths, though in prompts we have emphasized that GeoGPT can only use tools we defined and files it has obtained. In this way, designing such a protection mechanism can stabilize our framework to a certain degree. However, such a mechanism is only a compensation strategy, and investigating an effective solution to make LLMs operate reliably still needs more exploration.

\begin{longtable}{|p{14cm}|}
    \caption{An example to show the protection mechanism.}
    \label{table9}
\\

    \hline
    \textbf{Query}: I want to find hotels that locate meanwhile within a maximum distance of 500 meters from subway stations, and within neighboring areas of 800 meters from Hubei University.\\
    \hline
    \endfirsthead
    \hline
    \endhead

\hline
\multicolumn{1}{r}{\textit{Continued on next page}} \\
\endfoot

\hline
\endlastfoot
\textbf{Thought}: I need to find hotels that are within 500 meters of subway stations and 800 meters of Hubei University. I need to use the tools provided to obtain the necessary data and then use the Clip and Intersect tools to find the hotels that meet the criteria.\\
\textbf{Action}: Load\_Hotel\_Data\\
\textbf{Action Input}: None\\
\textbf{Observation}: I obtain the dataset of hotels, it is stored in ./data\_use/case3/hotel.shp\\
\\
\textbf{Thought}:I need to obtain the location of Hubei University and the subway stations.\\
\textbf{Action}: Obtain location\\
\textbf{Action Input}: Hubei University\\
\textbf{Observation}: I obtain the location of Hubei University, it is stored in ./data\_output/Hubei University.shp\\
\\
\textbf{Thought}:\\
\textbf{Action}: Obtain location\\
\textbf{Action Input}: Subway Stations\\
\textbf{Observation}: I obtain the location of Subway Stations, it is stored in ./data\_output/Subway Stations.shp\\
\\
\textbf{Thought}:\\
\textbf{Action}: Buffer\\
\textbf{Action Input}: Subway Stations, 500 \textcolor{red}{(1)}\\
\textbf{Observation}: An exception occurred using this tool, I need to check whether the input file exists, or whether the input contains excessive content, or whether the input format is correct, or whether the input file order is correct, or and whether I have selected the wrong tool\\
\\
\textbf{Thought}: I need to check whether the input file exists. \textcolor{red}{(1)}\\
\textbf{Action}: Load\_Subway\_Data\\
\textbf{Action Input}: None\\
\textbf{Observation}: I obtain the dataset of subway stations, it is stored in ./data\_use/case2/subway\_station.shp\\
\\
\textbf{Thought}: I need to use the Buffer tool to obtain the neighboring range around the subway stations.\\
\textbf{Action}: Buffer\\
\textbf{Action Input}: Subway Stations, 500\\
\textbf{Observation}: An exception occurred using this tool, I need to check whether the input file exists, or whether the input contains excessive content, or whether the input format is correct, or whether the input file order is correct, or and whether I have selected the wrong tool\\
\\
\textbf{Thought}: I need to check whether the input file order is correct.\\
\textbf{Action}: Buffer\\
\textbf{Action Input}: ./data\_use/case2/subway\_station.shp, 500\\
\textbf{Observation}: I obtain the neighboring area with a distance of 500.0 for ./data\_use/case2/subway\_station.shp, it is stored in ./data\_output/subway\_station\_neighborhood\_500.0.shp\\
\\
\textbf{Thought}: I need to use the Buffer tool to obtain the neighboring range around Hubei University.\\
\textbf{Action}: Buffer\\
\textbf{Action Input}: Hubei University, 800 \textcolor{red}{(2)}\\
\textbf{Observation}: An exception occurred using this tool, I need to check whether the input file exists, or whether the input contains excessive content, or whether the input format is correct, or whether the input file order is correct, or and whether I have selected the wrong tool \textcolor{red}{(2)}\\
\\
\textbf{Thought}: I need to check whether the input file exists.\\
\textbf{Action}: Load\_Hubei\_University\_Data \textcolor{red}{(3)}\\
\textbf{Action Input}: None\\
\textbf{Observation}: Load\_Hubei\_University\_Data is not a valid tool, try another one. \textcolor{red}{(3)}\\
\\
\textbf{Thought}:\\
\textbf{Action}: Obtain location\\
\textbf{Action Input}: Hubei University\\
\textbf{Observation}: I obtain the location of Hubei University, it is stored in ./data\_output/Hubei University.shp\\
\\
\textbf{Thought}:\\
\textbf{Action}: Buffer\\
\textbf{Action Input}: Hubei University, 800\\
\textbf{Observation}: An exception occurred using this tool, I need to check whether the input file exists, or whether the input contains excessive content, or whether the input format is correct, or whether the input file order is correct, or and whether I have selected the wrong tool\\
\\
\textbf{Thought}: I need to check whether the input file order is correct.\\
\textbf{Action}: Buffer\\
\textbf{Action Input}: ./data\_output/Hubei University.shp, 800\\
\textbf{Observation}: I obtain the neighboring area with a distance of 800.0 for ./data\_output/Hubei University.shp, it is stored in ./data\_output/Hubei University\_neighborhood\_800.0.shp\\
\\
\textbf{Thought}:\\
\textbf{Action}: Clip\\
\textbf{Action Input}: ./data\_use/case3/hotel.shp, ./data\_output/Hubei University\_neighborhood\_800.0.shp\\
\textbf{Observation}: In this step I obtain objects of ./data\_use/case3/hotel.shp within ./data\_output/Hubei University\_neighborhood\_800.0.shp, it is stored in ./data\_output/clip.shp\\
\\
\textbf{Thought}:\\
\textbf{Action}: Intersect\\
\textbf{Action Input}: ./data\_output/clip.shp,\\ ./data\_output/subway\_station\_neighborhood\_500.0.shp\\
\textbf{Observation}: In this step I obtain objects that meanwhile locate within ./data\_output/subway\_station\_neighborhood\_500.0.shp,./data\_output/clip.shp, it is stored in ./data\_output/intersect.shp\\
\\
\textbf{Final Answer}: The hotels that locate meanwhile within a maximum distance of 500 meters from subway stations, within neighboring areas of 800 meters from Hubei University are stored in ./data\_output/intersect.shp.\\

\hline

    \hline
\end{longtable}

\section{Conclusion and future work}
In this paper, we propose a framework called GeoGPT that can automatically solve some geospatial tasks with the instruction of only natural language. Our GeoGPT is designed with the Langchain framework, and we set the well-known GPT LLM gpt-3.5-turbo as the agent to control and call different GIS tools to tackle different problems. In this framework, classical GIS tools including Buffer, Clip, Intersect, and some professional tools like land use classification are defined to support the agent with the ability to solve geospatial tasks. Then, we conduct four representative case studies including geospatial data crawling, facility sitting, spatial query, and mapping to verify the effectiveness of our framework. From the case studies, GeoGPT can understand the demands of users merely based on non-professional descriptions, and then think, plan, and execute our defined GIS tools to output final effective results. Therefore, we think integrating the semantic understanding ability of LLMs and professional GIS tools provides a feasible way to prompt GIS to more people, and non-professional users can also get the help of GIS without a lot of learning cost. Moreover, we think the paradigm of “foundational plus professional” implied in GeoGPT also provides an effective way to develop next-generation GIS in this era of large foundation models.

However, there are still many challenges needed to be solved. First, how to align and harmonize professional knowledge and universal knowledge in LLMs is a critical problem. To interpret users’ demands and then call corresponding tools, LLMs should be equipped with the ability to translate non-professional descriptions into professional steps. Then, since spatial analysis in GIS should be accurate and reliable, the output of LLMs should be consistent and correct each time; while for some cartography mapping tasks, the color scheme of maps should have diversity. Therefore, how to balance the requirements of these tasks is also an important problem.

\section{References}
[1] D. J. Maguire, "An overview and definition of GIS," Geographical information systems: Principles and applications, vol. 1, no. 1, pp. 9-20. 1991.

\noindent [2] S. Fazal, GIS basics: New Age International, 2008.

\noindent [3] M. F. Goodchild, "GIS and modeling overview," GIS, spatial analysis, and modeling. ESRI Press, Redlands, pp. 1-18. 2005.

\noindent [4] A. G. Yeh, "Urban planning and GIS," Geographical information systems, vol. 2, no. 877-888, p. 1. 1999.

\noindent [5] M. Kahila-Tani, M. Kytta, and S. Geertman, "Does mapping improve public participation? Exploring the pros and cons of using public participation GIS in urban planning practices," Landscape Urban Plan., vol. 186, pp. 45-55. 2019.

\noindent [6] Y. Choi, J. Baek, and S. Park, "Review of GIS-based applications for mining: Planning, operation, and environmental management," Applied Sciences, vol. 10, no. 7, p. 2266. 2020.

\noindent [7] A. Singh, "Remote sensing and GIS applications for municipal waste management," J. Environ. Manage., vol. 243, pp. 22-29. 2019.

\noindent [8] F. Wang, "Why public health needs GIS: a methodological overview," Annals of GIS, vol. 26, no. 1, pp. 1-12. 2020.

\noindent [9] B. F. Khashoggi and A. Murad, "Issues of healthcare planning and GIS: a review," ISPRS International Journal of Geo-Information, vol. 9, no. 6, p. 352. 2020.

\noindent [10] M. A. Abdelrahman and S. Tahoun, "GIS model-builder based on comprehensive geostatistical approach to assess soil quality," Remote sensing Applications: society and Environment, vol. 13, pp. 204-214. 2019.

\noindent [11] P. Csáfordi, A. Pődör, J. Bug, and Z. Gribovszki, "Soil erosion analysis in a small forested catchment supported by ArcGIS Model Builder," Acta Silvatica et Lignaria Hungarica, vol. 8, pp. 39-55. 2012.

\noindent [12] E. Kasneci et al., "ChatGPT for good? On opportunities and challenges of large language models for education," Learning and Individual Differences, vol. 103, p. 102274. 2023.

\noindent [13] M. Sallam, "The utility of ChatGPT as an example of large language models in healthcare education, research and practice: Systematic review on the future perspectives and potential limitations," medRxiv, pp. 2022-2023. 2023.

\noindent [14] F. Tustumi, N. A. Andreollo, and J. E. D. Aguilar-Nascimento, "Future of the language models in healthcare: the role of chatGPT," ABCD. Arquivos Brasileiros de Cirurgia Digestiva (São Paulo), vol. 36, p. e1727. 2023.

\noindent [15] S. Hao et al., "Reasoning with language model is planning with world model," arXiv preprint arXiv:2305.14992. 2023.

\noindent [16] T. B. Richards, "Auto-gpt: An autonomous gpt-4 experiment,". 2023.

\noindent [17] C. Wu, S. Yin, W. Qi, X. Wang, Z. Tang, and N. Duan, "Visual chatgpt: Talking, drawing and editing with visual foundation models," arXiv preprint arXiv:2303.04671. 2023.

\noindent [18] J. Wang et al., "Review of Large Vision Models and Visual Prompt Engineering," arXiv preprint arXiv:2307.00855. 2023.

\noindent [19] A. M. Bran, S. Cox, A. D. White, and P. Schwaller, "ChemCrow: Augmenting large-language models with chemistry tools," arXiv preprint arXiv:2304.05376. 2023.

\noindent [20] H. Dai et al., "AD-AutoGPT: An Autonomous GPT for Alzheimer's Disease Infodemiology," arXiv preprint arXiv:2306.10095. 2023.

\noindent [21] J. Wei et al., "Emergent abilities of large language models," arXiv preprint arXiv:2206.07682. 2022.

\noindent [22] W. X. Zhao et al., "A survey of large language models," arXiv preprint arXiv:2303.18223. 2023.

\noindent [23] T. Brown et al., "Language models are few-shot learners," Advances in neural information processing systems, vol. 33, pp. 1877-1901. 2020.

\noindent [24] R. Openai, "Gpt-4 technical report," arXiv. 2023.

\noindent [25] A. Chowdhery et al., "Palm: Scaling language modeling with pathways," arXiv preprint arXiv:2204.02311. 2022.

\noindent [26] R. Anil et al., "Palm 2 technical report," arXiv preprint arXiv:2305.10403. 2023.

\noindent [27] H. Touvron et al., "Llama: Open and efficient foundation language models," arXiv preprint arXiv:2302.13971. 2023.

\noindent [28] Y. Tan et al., "Evaluation of ChatGPT as a question answering system for answering complex questions," arXiv preprint arXiv:2303.07992. 2023.

\noindent [29] A. Elhafsi, R. Sinha, C. Agia, E. Schmerling, I. Nesnas, and M. Pavone, "Semantic Anomaly Detection with Large Language Models," arXiv preprint arXiv:2305.11307. 2023.

\noindent [30] P. Mooney, W. Cui, B. Guan, and L. Juhász, "Towards Understanding the Spatial Literacy of ChatGPT--Taking a Geographic Information Systems (GIS) Exam,". 2023.

\noindent [31] M. Karabacak and K. Margetis, "Embracing Large Language Models for Medical Applications: Opportunities and Challenges," Cureus, vol. 15, no. 5. 2023.

\noindent [32] Y. Gao, T. Sheng, Y. Xiang, Y. Xiong, H. Wang, and J. Zhang, "Chat-rec: Towards interactive and explainable llms-augmented recommender system," arXiv preprint arXiv:2303.14524. 2023.

\noindent [33] R. Zhang et al., "Prompt, generate, then cache: Cascade of foundation models makes strong few-shot learners," Proceedings of the IEEE/CVF Conference on Computer Vision and Pattern Recognition, 2023, pp. 15211-15222.

\noindent [34] A. Ramesh, P. Dhariwal, A. Nichol, C. Chu, and M. Chen, "Hierarchical text-conditional image generation with clip latents," arXiv preprint arXiv:2204.06125. 2022.

\noindent [35] A. Ramesh et al., "Zero-shot text-to-image generation," International Conference on Machine Learning, PMLR, 2021, pp. 8821-8831.

\noindent [36] J. Roberts, T. Lüddecke, S. Das, K. Han, and S. Albanie, "GPT4GEO: How a Language Model Sees the World's Geography," arXiv preprint arXiv:2306.00020. 2023.

\noindent [37] G. Mai et al., "On the opportunities and challenges of foundation models for geospatial artificial intelligence," arXiv preprint arXiv:2304.06798. 2023.

\noindent [38] S. Mitrović, D. Andreoletti, and O. Ayoub, "Chatgpt or human? detect and explain. explaining decisions of machine learning model for detecting short chatgpt-generated text," arXiv preprint arXiv:2301.13852. 2023.

\noindent [39] S. Zheng, J. Huang, and K. C. Chang, "Why Does ChatGPT Fall Short in Answering Questions Faithfully?" arXiv preprint arXiv:2304.10513. 2023.

\noindent [40] G. Mialon et al., "Augmented language models: a survey," arXiv preprint arXiv:2302.07842. 2023.

\noindent [41] R. Nakano et al., "Webgpt: Browser-assisted question-answering with human feedback," arXiv preprint arXiv:2112.09332. 2021.

\noindent [42] J. Wei et al., "Chain of thought prompting elicits reasoning in large language models," arXiv preprint arXiv:2201.11903. 2022.

\noindent [43] T. Kojima, S. S. Gu, M. Reid, Y. Matsuo, and Y. Iwasawa, "Large language models are zero-shot reasoners," arXiv preprint arXiv:2205.11916. 2022.

\noindent [44] J. Wei et al., "Chain-of-thought prompting elicits reasoning in large language models," Advances in Neural Information Processing Systems, vol. 35, pp. 24824-24837. 2022.

\noindent [45] M. Ahn et al., "Do as i can, not as i say: Grounding language in robotic affordances," arXiv preprint arXiv:2204.01691. 2022.

\noindent [46] W. Huang et al., "Inner monologue: Embodied reasoning through planning with language models," arXiv preprint arXiv:2207.05608. 2022.

\noindent [47] S. Yao et al., "React: Synergizing reasoning and acting in language models," arXiv preprint arXiv:2210.03629. 2022.

\noindent [48] Y. Shen, K. Song, X. Tan, D. Li, W. Lu, and Y. Zhuang, "Hugginggpt: Solving ai tasks with chatgpt and its friends in huggingface," arXiv preprint arXiv:2303.17580. 2023.

\noindent [49] H. Chase, "LangChain," https://github.com/hwchase17/langchain. 2022.

\end{document}